\documentclass[runningheads]{llncs}

\usepackage{eccv}

\usepackage{eccvabbrv}

\usepackage[dvipsnames]{xcolor}
\usepackage{booktabs}
\usepackage{multirow}
\usepackage{multicol}
\usepackage{colortbl}
\usepackage[table]{xcolor}

\usepackage{multirow}
\usepackage{tabularx}

\usepackage[many]{tcolorbox}
\usepackage{algorithm}
\usepackage{xcolor}

\usepackage{tabularx}
\usepackage{float}
\usepackage[dvipsnames]{xcolor}
\usepackage{bm}
\usepackage[percent]{overpic}
\usepackage{makecell}
\usepackage{xspace}
\usepackage{wrapfig}
\usepackage{caption}

\newcommand{\best}{\cellcolor[HTML]{FFCCC9}}
\newcommand{\second}{\cellcolor[HTML]{FFE4CF}}
\newcommand{\third}{\cellcolor[HTML]{FFFFD4}}

\newcommand{\our}{\cellcolor{blue!5}}
\newcommand{\topa}{\colorbox[HTML]{FFCCC9}}
\newcommand{\topb}{\colorbox[HTML]{FFE4CF}}
\newcommand{\topc}{\colorbox[HTML]{FFFFD4}}

\newcolumntype{g}{>{\columncolor{blue!5}}c}
\usepackage[skip=3pt]{caption}
\usepackage{graphicx}
\usepackage{booktabs}

\usepackage[accsupp]{axessibility}  % Improves PDF readability for those with disabilities.

\usepackage{hyperref}

\usepackage{orcidlink}

\begin{document}

\title{PhysConvex: Physics-Informed Dynamic Convex Fields for Reconstruction and Simulation} 

\titlerunning{PhysConvex}

% Second Author\inst{2,3}\orcidlink{1111-2222-3333-4444} \and
% Third Author\inst{3}\orcidlink{2222--3333-4444-5555}}
\author{Dan Wang\textsuperscript{\rm 1*}, 
Xinrui Cui\textsuperscript{\rm 2*}, 
Serge Belongie\textsuperscript{\rm 3}, 
Ravi Ramamoorthi\textsuperscript{\rm 1}
}
\authorrunning{D.~Wang, X.~Cui, et al.}

% Springer Heidelberg, Tiergartenstr.~17, 69121 Heidelberg, Germany
\institute{\textsuperscript{\rm 1} UC San Diego %\\
    \textsuperscript{\rm 2} University of North Texas %\\
    \textsuperscript{\rm 3} University of Copenhagen \\
    {\small \textsuperscript{*}Equal Contribution \quad
    % \textsuperscript{\dag}Corresponding Author \quad
\href{https://danalaina.github.io/physconvex_webpage/}{Project Page}
}
}

\maketitle

\begin{center}
    \captionsetup{type=figure}
    \includegraphics[width=0.975\textwidth]{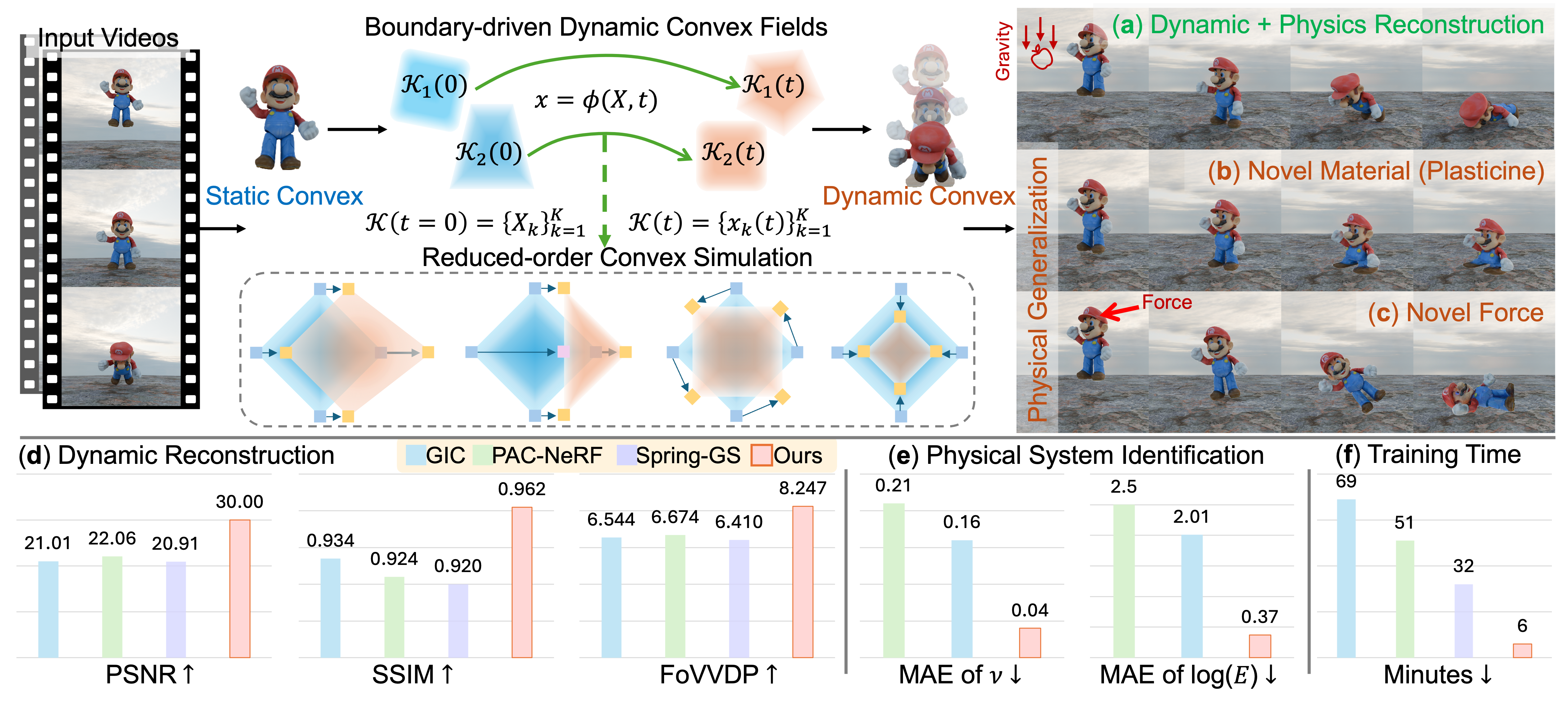}
    % \vspace{-0.2cm}
     \vspace{-2pt}
    \caption{PhysConvex introduces boundary-driven dynamic convex fields integrated with mesh-free reduced-order convex simulation for dynamic reconstruction, system identification(a), physical generalization(b,c). It recovers appearance, geometry, and physics from videos, improving dynamic and physical reconstruction(d,e), training efficiency(f).}
     \vspace{-8pt}
    \label{fig:teaser}
\end{center}%
%\vspace{-5pt}

 \vspace{-5pt}
\begin{abstract}
Reconstructing deformable objects from video requires a 4D representation that should preserve geometry, explain motion through physics, and generalize to future or new physical conditions. Existing dynamic NeRF and Gaussian methods achieve strong view synthesis, yet their voxel or ellipsoidal primitives are primarily designed for appearance rendering and typically driven by centers or predefined particle bindings, limiting physically meaningful non-uniform deformation and sharp boundary evolution. We present \textbf{PhysConvex}, a physics-informed dynamic convex field for video-based reconstruction, physical system identification, and simulation. PhysConvex represents a dynamic object as material-space deformable convex primitives whose boundary is advected by physical dynamics. The boundary-driven convex expresses non-uniform deformation and evolving active supports while serving simultaneously as a rendering element, deformation carrier, and physical support where mass, elastic response, forces, and contacts are evaluated. We further introduce a mesh-free reduced-order convex simulator in which neural skinning modes define physics-based deformation bases directly over deformable convex supports. Experiments show improved dynamic and physical reconstruction, efficiency, future prediction, and generalization to changed materials, forces, and boundary conditions. 
\vspace{-5pt}
\end{abstract}
\vspace{-5pt}

 \vspace{-5pt}
 \section{Introduction}
\label{sec:intro}
 \vspace{-5pt}

Videos capture not only how objects look, but also how they move, deform, and respond to the physical world. A dynamic 4D representation should preserve appearance, geometry, and physical behavior: it can be rendered from novel views, explained through physical properties, and simulated under new forces, contacts, or material conditions (Fig.~\ref{fig:teaser}). This is especially challenging for deformable objects, where appearance, geometry, and physical properties are entangled in video observations.
Classical physics-based simulation~\cite{diazzi2023constrained,li2020incremental,benchekroun2023FastComplemDynamics, Zhang:CompDynamics:2020,schneider2019poly} provides physically plausible dynamics, but typically assumes known geometry, meshing, and material parameters, limiting its real-world applicability where geometries and materials are often unknown.
Alternatively, purely neural approaches~\cite{wu20244d,luiten2024dynamic, duan20244d,lin2024gaussian, yang2024deformable, yang2023real, brooks2024video, tian2024visual} directly model dynamics through learned networks from data, disregarding underlying physical laws and thereby limiting generalization and interpretability. 

\begin{figure}[t]
    \centering
    \includegraphics[width=1\textwidth]{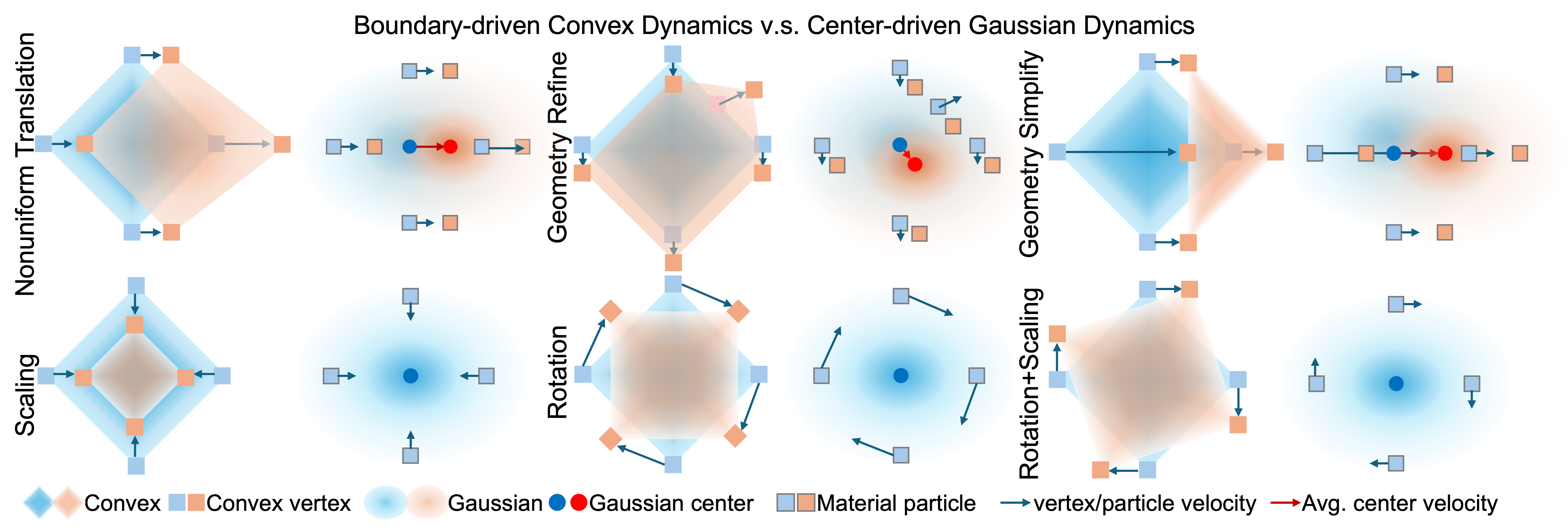} 
    \vspace{-5pt} 
    \caption{\footnotesize  
    Center-driven Gaussian dynamics have limited flexibility for non-uniform deformation and evolving boundaries, while boundary-driven convex dynamics enable spatially adaptive vertex advection, explicit surface evolution, topological change.}
    \label{fig:convex}
  \vspace{-15pt}
 %Approximate convex decomposition~\cite{deng2020cvxnet} vs. Gaussian decomposition~\cite{Genova_2019_ICCV}.
\end{figure}

Recent advances~\cite{li2023pac,xie2023physgaussian,zhong2024springgaus,jiang2025phystwin,chen2025vid2sim} bridge this gap by integrating neural radiance fields with differentiable physical simulation. 
These methods make important progress toward video-based physics and dynamic reconstruction, but most inherit a mismatch between rendering and simulation. 
They still rely on primitives (voxels in NeRF-based methods~\cite{li2023pac}, ellipsoids in Gaussian-based methods~\cite{zhong2024springgaus,cai2024gic,chen2025vid2sim}) that were designed for static appearance rendering efficiency rather than dynamic physical deformation, resulting in several limitations in dynamic scenarios (Fig.~\ref{fig:convex}):
First, in hybrid Eulerian-Lagrangian frameworks ~\cite{xie2023physgaussian}, center-driven dynamics updates primitives using central particles or pooled neighbors, lacking spatial sensitivity to non-uniform deformation; 
Second, predefined particle-primitive bindings ~\cite{zhong2024springgaus,Cao_2024_NeuMA} neglect material and geometric relations, resulting in inaccurate dynamic coupling; 
Third, existing NeRF and Gaussian representations integrated with grid-based MPM simulation~\cite{jiang2016material} struggle to capture hard or evolving boundaries, hindering physically meaningful surface deformation;
Finally, approximating Gaussian decomposition by symmetric, rounded ellipsoids limits the modeling of anisotropic geometric structures or nonlinear materials, while also producing poor spatial coverage near planar or angular regions due to inherent sphere-packing and vacuum-gap issues~\cite{conway2013sphere, gross2007point}.

Our key insight is that \emph{the rendering primitive should also be the material support for physical deformation and simulation}.
Based on this principle, we introduce \textbf{PhysConvex}, a physics-informed dynamic convex field for video-based reconstruction, physical system identification, and future-state simulation (Fig.~\ref{fig:teaser}). PhysConvex represents a dynamic object as a set of material-space deformable 4D convex primitives. Unlike a Gaussian primitive, which is mainly controlled by a center and covariance, each PhysConvex primitive is defined by a deformable convex hull governed by continuum mechanics to model dynamics.

We introduce a \textbf{boundary-driven dynamic convex representation} (Sec. \ref{sec:boundarysconvex}) that gives each primitive spatially adaptive deformation capacity, rather than forcing it to move as a center-driven kernel (Fig.~\ref{fig:convex}). We represent convex boundary deformation explicitly as vertex dynamics of convex hulls, or implicitly as the surface dynamics of half-space support functions: 
\textit{Vertex dynamics} advect individual convex vertices under Newtonian motion, providing spatial sensitivity and flexibility to non-uniform deformation; \textit{Surface dynamics} captures evolving edges and boundaries, allowing physically meaningful surface deformation; \textit{Convex dynamics} enable polyhedral structural modification, enhancing geometric expressiveness and adaptability for simulating anisotropic materials. 

To integrate convex primitives into physics-based dynamics, we develop a \textbf{mesh-free reduced-order convex simulation} (Sec. \ref{sec:reducedsconvex}).
Dynamic convexes are advected under continuum mechanics, using a continuous implicit neural skinning field as physics-informed, shape- and material-aware deformation modes, with time-varying handle transformations as reduced degrees of freedom under Newtonian dynamics~\cite{modi2024Simplicits}.
The neural skinning field captures low-dimensional physically plausible deformation modes directly over the material-space convex supports without requiring a mesh, background grid, or manually specified primitive-particle correspondence.
PhysConvex therefore learns a physically grounded deformable convex field for dynamics and physics reconstruction via \textit{differentiable rendering and simulation} (Sect. \ref{sec:rendersimulator}).
Once optimized, the same representation can be rendered from novel views, rolled forward for future prediction, or simulated under changed material, force, and boundary conditions.

In PhysConvex, our dynamic 4D convex is simultaneously a rendering element, a deformation carrier, and a physical support. 
For rendering, the deformed boundary induces a smooth convex occupancy that can be splatted differentiably. For dynamics, the same material-space support allows each primitive to translate, rotate, shear, stretch, and change its active supporting faces over time, and defines where mass, elastic response, forces, and contacts are evaluated.
This shared representation avoids predefined bindings between rendering primitives and simulation particles, and provides compact coverage of volumetric regions, flat surfaces, and angular structures. It extends geometric advantages of static convex splatting~\cite{Held20243DConvex} to deformable, physically grounded dynamic reconstruction.

Our contributions are:
\textbf{(1)} We propose \textbf{PhysConvex}, a dynamic convex field representation for video-based reconstruction and physically grounded simulation of deformable objects.
\textbf{(2) }We introduce a \textbf{boundary-driven dynamic convex representation} whose vertex and surface boundaries, rather than only its center, are advected by a reduced deformation field. This enables non-uniform deformation, evolving active supports, and sharper geometric structures than center-driven Gaussian or voxel primitives.
\textbf{(3) }We formulate a \textbf{mesh-free reduced-order convex simulation} framework in which neural skinning modes define a physical deformation basis over material-space convex supports. The same convex primitives are used for rendering, deformation, mass, elastic energy, external forces, and contact, avoiding predefined particle-primitive bindings.
\textbf{(4) }We demonstrate that PhysConvex improves dynamic reconstruction and physical system identification with better training efficiency than physics-integrated NeRF and Gaussian baselines, while supporting efficient future-state prediction and generalization to changed materials, forces, and boundary conditions.

 \section{Related Work}
\label{sec:related}
\textbf{Physics-Based Simulation.}
Classical physics simulation relies on mesh-based FEM~\cite{cutler2002procedural,bargteil2007finite,bouaziz2023projective,schneider2019poly}, which requires explicit meshing and linear shape functions, making it inflexible for complex or evolving geometries. Mesh-free methods such as MPM~\cite{jiang2016material,klar2016drucker,hu2019difftaichi,yue2015continuum} and SPH~\cite{desbrun1996smoothed,kugelstadt2021fast,peer2018implicit} alleviate meshing but still depend on background grids or neighborhood connectivity, limiting efficiency and stability for highly nonlinear dynamics. Recent efforts, such as Spring-Gaus~\cite{zhong2024springgaus}, employ mass-spring systems for elastic behavior, but remain restricted in generalizing to diverse materials. Model Order Reduction approaches~\cite{Zhang:CompDynamics:2020,benchekroun2023FastComplemDynamics,chang2023licrom,modi2024Simplicits,chen2022crom} offer efficient, mesh-free alternatives by learning reduced-order subspaces that enable accurate, real-time simulation of complex deformations. Inspired by these advances, our method develops a reduced-order convex simulation model that integrates vertex- and surface-based dynamics with differentiable MOR simulation and reconstruction for geometry, appearance, and physical properties.

\smallskip
\noindent\textbf{Video-based Physical Reconstruction and Simulation.}
 % \vspace{-5pt}
Integrating physical simulation with neural 3D reconstruction has gained traction in works~\cite{li2023pac,xie2023physgaussian,zhong2024springgaus,jiang2025phystwin,chen2025vid2sim}, integrating mesh-free simulation with NeRF or 3D Gaussian Splatting. 
However, most existing frameworks rely on neural primitives, such as voxels in NeRF~\cite{li2023pac} or ellipsoids in 3D Gaussian Splatting~\cite{xie2023physgaussian,zhong2024springgaus,jiang2025phystwin}, combined with physics-based simulation. However, these primitives are primarily optimized for efficient rendering rather than modeling physically consistent deformation, resulting in several key limitations: lack of spatial deformation flexibility within primitives; difficulty modeling sharp or evolving boundaries; restricted shape expressiveness due to fixed-shape Gaussian or voxel kernels; and inaccurate binding between Eulerian rendering primitives and Lagrangian dynamics particles.
To address this gap, we introduce PhysConvex, a unified framework for visual rendering and physical simulation. Inspired by 3DCS~\cite{Held20243DConvex}, PhysConvex presents a physical dynamic convex representation to reconstruct and simulate radiance fields governed by continuum mechanics.

 \vspace{-5pt}
\section{Preliminaries}
\vspace{-5pt}
\textit{Continuum Mechanics.}
 \label{sec:continuum}
Continuum mechanics describes the motion of deformable objects through a time-dependent deformation map $\mathbf{x} = \boldsymbol{\phi}(\mathbf{X}, t)$, which maps material coordinates $\mathbf{X}$ in the undeformed reference domain $\Omega^0$ to world coordinates $\mathbf{x}$ in the deformed configuration $\Omega^t$ at time $t$. The evolution of $\boldsymbol{\phi}$ is governed by the conservation of mass and momentum:
\begin{equation}
\begin{aligned}
  & \int_{\Omega^t} \rho(\mathbf{x}, t) \equiv \int_{\Omega^0} \rho(\boldsymbol{\phi}^{-1}(\mathbf{x}, t), 0),\quad
 \rho(\mathbf{x}, t) \dot{\mathbf v}(\mathbf{x}, t) = \nabla \cdot \boldsymbol{\sigma}(\mathbf{x}, t) + \mathbf{f}^{\text{ext}},
\end{aligned}
\end{equation}
where $\rho(\mathbf{x}, t)$ is the density field, $\dot{\mathbf{v}}(\mathbf{x}, t) = \frac{\partial^2 \boldsymbol{\phi}}{\partial t^2}$ is the acceleration, and $\mathbf{f}^{\text{ext}}$ denotes the external force. The internal force is given by the Cauchy stress tensor $\boldsymbol{\sigma}(\mathbf{x}, t)$, which encodes local deformation behavior. The Jacobian of the deformation map, $\mathbf{F}(\mathbf{X}, t) = \nabla_{\mathbf{X}} \boldsymbol{\phi}(\mathbf{X}, t)$, is the deformation gradient, representing local stretch, rotation, and shear, and is central to elasticity. The constitutive model defines the relationship between $\mathbf{F}$, the strain energy density function $\Psi(\mathbf{F})$, and the stress tensor $\boldsymbol{\sigma}$, parameterized by Young’s modulus $E$ and Poisson’s ratio $\nu$.

 % \vspace{-10pt}
\textit{Model Order Reduction.}
\label{sec:mor}
Model order reduction (MOR) provides a compact, mesh-free alternative with efficient numerical integration and accurate modeling of nonlinear deformation. In MOR, the deformation map is parameterized as $\mathbf{x} = \boldsymbol{\phi}(\mathbf{X}, \mathbf{z}(t))$, where $\mathbf{z}(t)$ is a time-varying low-dimensional vector of reduced degrees of freedom (DOFs). Using implicit time integration, the simulation is formulated as an optimization problem:
\begin{equation}
    \mathbf{z}_{t+1} = \arg\min_{\mathbf{z}} 
    \frac{1}{2}\|\mathbf{z} - \tilde{\mathbf{z}}_{t}\|_{\mathbf{M}}^2 
    + \Delta t^2 \mathbf{E}_{\text{potential}}(\mathbf{z}),
    \label{eq:time_stepping}
\end{equation}
where $\mathbf{E}_{\text{potential}}(\mathbf{z})$ denotes the total potential energy of the simulated object from internal and external forces, $\Delta t$ is the timestep, $\tilde{\mathbf{z}}_{t}$ is the first-order predictor, and $\|\cdot\|_{\mathbf{M}}^2$ is the mass-matrix-weighted norm.

 % \vspace{-10pt}
\textit{3D Convex Splatting.}
\label{sec:conv}
3D Convex Splatting (3DCS)~\cite{Held20243DConvex} represents radiance fields using smooth convex primitives, offering greater geometric expressiveness than Gaussians and enabling efficient modeling of scenes with sharp edges and dense volumes using fewer elements. A 3D convex is defined as the convex hull of a point set \mbox{$\mathcal{K} = \{\mathbf{x}_k\}_{k=1}^K$}, where the points $\{\mathbf{x}_k\}$ define the hull of the shape, but are not necessarily explicit polyhedral vertices. A convex radiance field consists of $N$ convex polytopes \mbox{$\mathcal{C} = \{\mathcal{K}_n, \mathbf{c}_n, \sigma_n, \alpha_n, \beta_n\}_{n=1}^N$}, where $\mathbf{c}_n$ and $\sigma_n$ denote color and opacity, and $\alpha_n$ and $\beta_n$ control smoothness and sharpness, respectively. Here, $\alpha$ governs the softness of vertices and edges, and $\beta$ modulates the radiance transition from diffuse to dense regions.

%==========================================
 \vspace{-5pt}
\section{Method}
\label{sec:method}
 \vspace{-5pt}
 Given posed videos of a deformable object, our goal is to reconstruct a dynamic 4D representation that can be rendered from novel views, explained by physical parameters, and simulated beyond the observed sequence.
PhysConvex is built on a design principle: the rendering primitive should also be the material support for deformation and simulation. Instead of attaching visual primitives to external particles or moving each primitive only by its center, we represent the dynamic object as material-space deformable convex primitives whose boundaries are advected by a reduced physical deformation model.

%============================================

\begin{figure*}[t]
    \centering
    \includegraphics[width=1\textwidth]{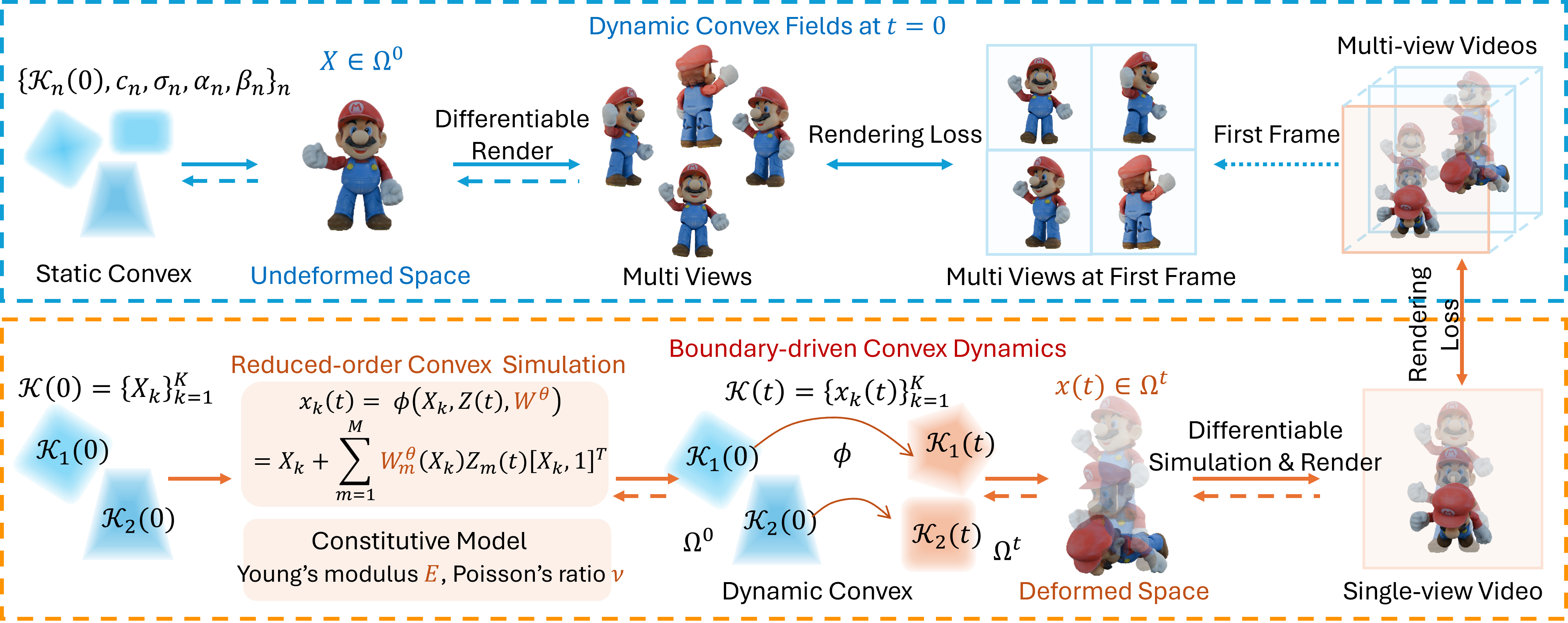} 
    \vspace{-10pt} 
   \caption{\footnotesize PhysConvex proposes differentiable boundary-driven dynamic convex field (Sec. \ref{sec:boundarysconvex}) and mesh-free reduced-order convex simulation (Sec. \ref{sec:reducedsconvex}), jointly reconstructing geometry, appearance, and physical properties of dynamic objects (Sec. \ref{sec:rendersimulator}).} 
   \vspace{-5pt}
    \label{fig:architecture}
  \vspace{-10pt}
\end{figure*}

\textit{Method Overview.}
PhysConvex introduces boundary-driven dynamic convex fields (Sec. \ref{sec:boundarysconvex}) integrated with physical reduced-order convex simulation (Sec. \ref{sec:reducedsconvex}) to achieve differentiable physically-consistent 3D reconstruction and dynamic modeling (Sec. \ref{sec:rendersimulator}). PhysConvex is trained by enforcing rendered pixels to match the observed video in two stages (Fig.~\ref{fig:architecture}): (1) An undeformed convex field is reconstructed over the first multi-view frame; (2) The reduced-order convex simulation advects our boundary-driven dynamic convex field under continuum mechanics, and physical properties can be optimized through joint differential simulation and rendering using single-view video supervision.

 \vspace{-8pt}
\subsection{Boundary-driven Dynamic Convex Fields}
\label{sec:boundarysconvex} 
\vspace{-5pt}
A dynamic object can be approximated as a collection of time-varying convex polytopes, forming a piecewise representation of dynamic geometry. By extending with an additional time variable $t\in\mathbb{R}^+$, we parameterize our dynamic convex field as a set of time-dependent $N$ convex polytopes:
\begin{equation}
\mathcal{C}(t) = \{\mathcal{K}_n(t), \mathbf{c}_n, \sigma_n, \alpha_n, \beta_n\}_{n=1}^N,
\end{equation}
where $\mathcal{K}_n(t)$ denotes the $n$-th dynamic convex at time $t$, and $\{\mathbf{c}_n, \sigma_n, \alpha_n, \beta_n\}$ are its time-invariant color, opacity, smoothness, and sharpness parameters. 

We represent boundary-driven convex deformation explicitly as vertex dynamics of convex hulls, or implicitly as surface dynamics of half-space support functions, enabling spatially adaptive vertex deformation, physically meaningful surface evolution, and geometry modification through polyhedral structures (Fig.~\ref{fig:convex}). Explicit vertex dynamics naturally support physical simulation, while implicit surface dynamics enable differentiable rendering.

\vspace{-7pt}
\subsection{Physical Reduced-order Convex Simulation}
\label{sec:reducedsconvex} 
\vspace{-5pt}

\textbf{Convex Vertex Dynamics.}
\label{sec:vertex}
 % \vspace{-2pt}
A dynamic convex is defined as a 3D time-varying point set $\mathcal{K}(t) = \{\mathbf{x}_k(t)\}_{k=1}^K$ whose convex hull represents the primitive shape. 
Here, $K$ is an upper bound on shape capacity rather than a fixed number of active hull vertices. 
During simulation, these points evolve under physical dynamics, allowing each convex primitive to adapt its shape, anisotropy, and boundary structure, thereby supporting spatially flexible, non-uniform deformation. Even with a constant K, the effective convex geometry can adapt through changes in active hull supports, enabling implicit refinement or simplification (Fig.~\ref{fig:convex}.a-2,3).
The undeformed convex hull is denoted as $\mathcal{K}(t=0) = \{\mathbf{X}_k\}_{k=1}^K$.

To advect dynamic convex fields, inspired by reduced-order simulation \cite{modi2024Simplicits}, we model the vertex dynamics $\mathbf{x}_k(t)$ using spatially varying neural skinning eigenmodes $W^\theta$ as reduced deformation bases, 
\begin{equation}
    \mathbf  \phi(\mathbf X_k,\mathbf Z(t), W^\theta)=\mathbf X_k+\sum_{m=1}^{M}W_m^\theta(\mathbf X_k)\mathbf Z_m(t)[\mathbf X_k,1]^T,
\end{equation}
where skinning eigenmodes $W^\theta(\mathbf X)\in\mathbb{R}^3\to\mathbb{R}^M$ represent physics-informed deformation bases encoding both shape and material awareness, and $M$ is the number of reduced DOFs. The reduced bases $W^\theta(\mathbf X)$ are represented as a small MLP-based neural field parameterized by $\theta$, enabling smooth, mesh-free discretization for complex geometries and heterogeneous materials. The neural skinning field $W^\theta(\mathbf X)$ is trained to learn physically significant modes by minimizing: 
\begin{equation}
\begin{aligned}
    \min_{\theta}\; &\lambda_{\text{ela}}\int_{\mathbb{R}^3} \varphi(\mathbf{X})\Psi(\phi_{}(\mathbf X,\mathbf Z, W^\theta))\,d\mathbf{X} \\
    &{}+\lambda_{\text{ort}}\sum_{i=1}^{M}\sum_{j=1}^{M}\int_{\mathbb{R}^3} \varphi(\mathbf{X})(W^{\theta}_iW^{\theta}_j-\delta_{ij})^2\,d\mathbf{X},
    \label{eq:EigenmodeField}
\end{aligned}
\end{equation}
where $\varphi(\mathbf{X})$ is a density function, $\Psi(\cdot)$ is the strain energy density function, and $\delta_{ij}$ is the Kronecker delta. 
$W^\theta(\mathbf X)$ is optimized in a data-free manner via elastic energy minimization (the first term in Eq.~\ref{eq:EigenmodeField}) and orthogonality regularization (the second term) with randomly sampled ($\mathbf{X},\mathbf{Z}$) pairs. Physically implausible deformations incur high elastic energy, while orthogonality constraint prevents collapse into trivial rigid modes. These objectives encourage physically plausible, low-energy deformation modes rather than memorization of object-specific trajectories.
It enables $W^\theta(\mathbf X)$ to encode low-dimensional, physically consistent deformation behaviors that are material- and shape-aware, thereby achieving mesh-free model reduction without requiring explicit geometric scaffolding.

During the simulation, $\mathbf Z(t)\in\mathbb{R}^{3\times4\times M}$  represents time-varying reduced DOFs as handle transformation matrices, which are flattened into $\mathbf z(t)=\text{flat}(\mathbf Z(t))\in\mathbb{R}^{12\times M}$. $\mathbf Z(t)$ is initialized to zero to ensure that $\mathbf{x}_k(0) = \mathbf{X}_k$. At each discrete time step, $\mathbf{Z}(t)$ is updated via implicit time integration using the incremental potential formulation (Eq.~\ref{eq:time_stepping} with an inertia term and a potential energy term) and solved through a standard Newton-based solver~\cite{li2020incremental,bonnans2006numerical}. This process advances the dynamic convex field under physically consistent reduced-order dynamics.

 % \vspace{-2pt}
\textbf{Convex Surface Dynamics. }
 % \vspace{-2pt}
Given a dynamic convex point set, the convex hull can equivalently be expressed by $H$ time-varying planes $\mathcal{H}(t)=\{(\mathbf{n}_h(t),d_h(t))\}_{h=1}^H$, where $\mathbf{n}_h(t)$ and $d_h(t)$ denote the normal and offset of the $h$-th plane, respectively. The signed distance of a point $\mathbf{x}$ to the $h$-th time-dependent plane is:
\begin{equation}
f_h(\mathbf{x},t) = \mathbf{n}_h(t)\!\cdot\!\mathbf{x} + d_h(t).
\end{equation}

Following the static indicator formulation in CvxNet~\cite{deng2020cvxnet}, we define $\mathcal{O}(\mathbf{x},t): \mathbb{R}^3 \rightarrow [0, 1]$ as the time-dependent occupancy function for each dynamic convex:
\begin{equation}
  \begin{aligned}
   \mathcal{O}(\mathbf{x},t)
   = \text{Sigmoid}\!\left(-\beta\,\tau(\mathbf{x},t)\right),\quad
   \tau(\mathbf{x},t)
   = \log\!\left(\sum_{h=1}^H \exp(\alpha f_h(\mathbf{x},t))\right),
  \end{aligned}
\end{equation}
where $\alpha$ and $\beta$ control convex smoothness and sharpness, respectively, and $\tau(\mathbf{x},t)$ approximates the signed distance function of a dynamic convex hull.

To render deformed convex kernels, we extend the splatting from 3DCS ~\cite{Held20243DConvex} to handle dynamic scenes. PhysConvex projects 3D time-varying points to 2D, constructs their dynamic convex hulls, and rasterizes them using $\alpha$-blending. Given camera pose and $N$ dynamic convexes, the time-varying pixel color at location $\mathbf{q}$ and time $t$ is:
\begin{equation}
\mathbf{C}(\mathbf{q},t) = \sum_{i=1}^N \mathbf{c}_i \sigma_i \mathcal{O}_i(\mathbf{q},t) 
\prod_{j=1}^{i-1} \big(1 - \sigma_j \mathcal{O}_j(\mathbf{q},t)\big),
\end{equation}
where $\mathbf{c}_i$ and $\sigma_i$ denote the color and opacity of the $i$-th dynamic convex.

\vspace{-5pt}
\subsection{Differentiable Reconstruction and Simulation}
\label{sec:rendersimulator}
\vspace{-5pt}
Given posed multi-view videos of a dynamic scene, we reconstruct the dynamic object’s appearance, geometry, and physical properties through a joint differentiable reconstruction and simulation, as shown in Fig.~\ref{fig:architecture}.

  % \vspace{-2pt}
\textbf{Appearance and Geometry Optimization. }
 % \vspace{-2pt}
To recover the undeformed convex field $\mathcal{C}(0)$, we optimize from the first multi-view frames of the input videos by minimizing:
\begin{equation}
\begin{aligned}
\mathcal{C}^*(0) 
&= \arg\min_{\{\mathcal{K}_n(0),\mathbf{c}_n,\sigma_n,\alpha_n,\beta_n\}} (1-\lambda)\mathcal{L}_1 + \lambda \mathcal{L}_{\text{D-SSIM}} + \beta \mathcal{L}_{\text{m}},
\end{aligned}
\end{equation}
where $\mathcal{L}_1$ and $\mathcal{L}_{\text{D-SSIM}}$ are image reconstruction losses, and $\mathcal{L}_{\text{m}}$ regularizes the number of convexes following \cite{Held20243DConvex}. This optimization yields the undeformed convex field $\mathcal{C}^*(0)=\{\mathcal{K}_n^*(0), \mathbf{c}_n^*, \sigma_n^*,\alpha_n^*,\beta_n^*\}_{n=1}^N$ in material space $\Omega^0$.

 % \vspace{-2pt}
\textbf{Physical Model Optimization.}
 % \vspace{-2pt}
To recover physical properties, we optimize the physical parameters $\{E, \nu\}$ with video supervision:
\begin{equation}
\begin{aligned}
&E^*, \nu^*
= \arg\min_{E,\nu} \frac{1}{SP}\sum_{p=1}^{P}\sum_{s=1}^{S}
\|\mathbf{C}(\mathbf{q}_p,t_s) - \mathbf{C}_{\text{gt}}(\mathbf{q}_p,t_s)\|_2^2,
\end{aligned}
\end{equation}
where $\mathbf{C}$ and $\mathbf{C}_{\text{gt}}$ denote the rendered and ground-truth pixel colors at each time step. Since both simulation and rendering are differentiable, physical and material parameters can be optimized end-to-end from video observations.

To equip our model with generalizable physical understanding, we adopt a feed-forward physical system identification framework inspired by \cite{chen2025vid2sim}. Given a single front-view video, a pretrained video Vision Transformer \cite{tong2022videomae} extracts motion cues and physical knowledge prior from large-scale video data. Small MLP regression heads then estimate $\{E, \nu, \theta\}$, providing an initial coarse prediction that is subsequently refined through differentiable simulation to better match reference videos.
In summary, PhysConvex jointly reconstructs geometry, appearance, and physical properties of deformable objects by unifying physical boundary-driven dynamic convex representation, differentiable reduced-order convex simulation, and rendering. 

\vspace{-5pt}
 \section{Results}
\label{sec:expt}
\vspace{-5pt}
In this section, we evaluate PhysConvex on four tasks:  physical system identification (Sec. \ref{sec:system}), dynamic reconstruction (Sec. \ref{sec:dynamic}), physical generalization capability (Sec. \ref{sec:generalization}), future state prediction (Sec. \ref{sec:future}). We also provide efficiency comparison (Sec. \ref{sec:time}), ablation study (Sec. \ref{sec:ablation}), real-world evaluation (Sec. \ref{sec:realdata}).

\textit{Google Scanned Object Dataset.}
Following \cite{chen2025vid2sim}, we use 12 high-quality 3D meshes from Google Scanned Objects (GSO) \cite{downs2022google}, featuring complex geometries and detailed textures. Objects are simulated using FEM to obtain physically accurate reference dynamics. The animations are rendered from 12 viewpoints at a resolution of $448\times448$ for 24 frames. The first 16 frames are used as observations, and the remaining 8 frames serve as ground-truth references for future state prediction. 
Following \cite{chen2025vid2sim}, Young’s modulus and Poisson’s ratio are randomly sampled from $E\in [10^4,10^6],\nu\in [0.2,0.5]$.

\textit{Real-world Dataset.}
We evaluate PhysConvex on real-world data following SpringGaus~\cite{zhong2024springgaus}. Real-world dataset includes both static scenes and dynamic multi-view videos.
Static scenes: Each object is placed on a table and captured with 50–70 multi-view images from the upper hemisphere. Camera poses are recovered using COLMAP~\cite{schonberger2016pixelwise,schonberger2016structure}.
Dynamic multi-view videos: Synchronized RGB videos are captured from three viewpoints at a resolution of $1920\times1080$. Intrinsic and extrinsic parameters are calibrated using a checkerboard.
Object masks for all scenes are obtained using the Segment Anything Model (SAM)~\cite{kirillov2023segment}.

\textit{Baselines. }
(1) \textit{PAC-NeRF} \cite{li2023pac} jointly reconstructed dynamic NeRFs and a simulatable model using MPM \cite{jiang2016material}.
(2) \textit{Spring-GS} \cite{zhong2024springgaus} models elastic objects with spring-mass 3D Gaussians and learnable spring stiffness parameters, limited to elastic dynamics.
(3) \textit{GIC} \cite{cai2024gic} integrates 3D Gaussians with MPM simulation, enhancing shape reconstruction via a motion-factorized network and coarse-to-fine filling strategy.
(4) \textit{Vid2Sim} \cite{chen2025vid2sim} is a feed-forward Gaussian-based model with reduced-order simulation by utilizing pre-trained VideoMAE \cite{tong2022videomae} and Large Multi-view Gaussian Model (LGM)~\cite{tang2024lgm} as backbones for reconstructing physical parameters and 3D Gaussians (opacity, RGB, $\mu$, $\Sigma$). 

\textit{Metrics.}
For \textit{physical parameters estimation}, we report Mean Absolute Error (MAE) of $\log(E)$ and $\nu$ under Neo-Hookean model (excluding Spring-GS). 
For \textit{dynamic reconstruction and future prediction}, we measure PSNR\cite{hore2010image}, SSIM\cite{wang2004image}, and video perceptual loss (FoVVDP) \cite{mantiuk2021fovvideovdp}. 

\textit{Implementation details. }
Each convex shape is initialized with $K = 6$ points and spherical harmonics of degree 3 (69 parameters per convex shape). We first train 3D convexes in the undeformed space, and then train the neural skinning network using the data-free method.
We then jointly optimize ${E, \nu, \theta}$ for 400 iterations with learning rates ${5\times10^{-3}, 1\times10^{-3}, 5\times10^{-7}}$ using the Adam optimizer~\cite{adam2014method}. Simulation uses $M=10$ control handles as reduced DOFs. The average training time cost per scene is about 6 min on a single Nvidia RTX A6000 (Details in Sec. \ref{sec:time}). Additional results for varying $K$ and $M$ appear in Sec. \ref{sec:ablation}. More details can be found in the Supplementary.

\vspace{-5pt}
\subsection{Physical System Identification}
 \label{sec:system}
\vspace{-5pt}
Following \cite{li2023pac,cai2024gic,chen2025vid2sim}, we assessed our method and all baselines on physical system identification against 12 GSO test cases.  As shown in Tab.~\ref{tab:physics}, PhysConvex achieves the lowest error in most cases and remains competitive in the rest, validating its accuracy in recovering physical properties.
\vspace{-5pt}
\begin{table*}[ht]
\vspace{-5pt}
\caption{\label{tab:physics} Comparison of Mean Absolute Error on Physical System Identification.}
\vspace{-5pt}
\resizebox{\textwidth}{!}{
\begin{tabular}{c|c|cccccccccccc|c}
\hline
& methods  & backpack & bell & blocks & bus  & cream & elephant & grandpa & leather & lion & mario & sofa & turtle & Mean Error $\downarrow$ \\ \hline
\multirow{4}{*}{$\log(E)$} & PAC-NeRF & 3.28     & 1.08 & 4.02   & 3.30 & 3.22  & 3.05     & 2.99    & 1.20    & 2.34 & 3.37  & 0.20 & 1.94   & 2.50       \\
                        
& Vid2Sim  &  0.69     & \textcolor{blue}{0.15} &  0.54   &\textcolor{blue}{ 0.26} &   0.95  &\textcolor{blue}{ 0.18}     & 1.07    &   0.73    &   0.48 & \textcolor{blue}{0.50}  &   0.18 & \textcolor{blue}{0.44}   &   0.51       \\
& GIC      & 1.16     & 2.87 & 2.12   & 1.93 & 2.13  & 1.53     & \textcolor{blue}{0.42}    & 3.45    & 2.85 & 1.82  & 0.65 & 3.18   & 2.01       \\
\rowcolor{blue!5} \cellcolor{white} & \textbf{Ours} & \textcolor{blue}{0.32}     &  0.34 & \textcolor{blue}{0.10}   &   0.28 &\textcolor{blue}{ 0.34}  &   0.60     &   0.97    &\textcolor{blue}{ 0.18 }   & \textcolor{blue}{0.23} &   0.54  &\textcolor{blue}{ 0.01} &  0.56   & \textcolor{blue}{0.37}       \\ 
                        \hline

\multirow{4}{*}{$\nu$}      & PAC-NeRF & 0.21     & 0.23 & 0.33   & 0.16 & 0.12  &   0.06     & 0.36    & 0.26    & 0.14 & 0.33  & 0.30 & 0.01   & 0.21       \\

& Vid2Sim  &   0.10     &   0.10 &   0.07   &   0.06 & 0.11  & \textcolor{blue}{0.05}     & \textcolor{blue}{0.02}    & 0.06    & 0.05 & \textcolor{blue}{0.06}  & 0.08 &   0.02   &   0.06       \\
& GIC      & 0.11     & 0.24 & 0.29   & 0.18 &   0.08  & 0.24     & 0.26    & \textcolor{blue}{0.02}    & 0.14 & 0.22  &\textcolor{blue}{ 0.01} & 0.08   & 0.16       \\                   
\rowcolor{blue!5} \cellcolor{white} & \textbf{Ours} & \textcolor{blue}{0.06}     & \textcolor{blue}{0.09} & \textcolor{blue}{0.02}   & \textcolor{blue}{0.02} & \textcolor{blue}{0.05}  & 0.08     &   0.03    & \textcolor{blue}{0.02}    & \textcolor{blue}{0.03} &   0.12  &   0.02 & \textcolor{blue}{0.01}   & \textcolor{blue}{0.04}       \\ \hline

\end{tabular}}
\end{table*}
\vspace{-5pt}

\vspace{-5pt}
\begin{figure*}[ht]
    \centering
    \includegraphics[width=1\textwidth]{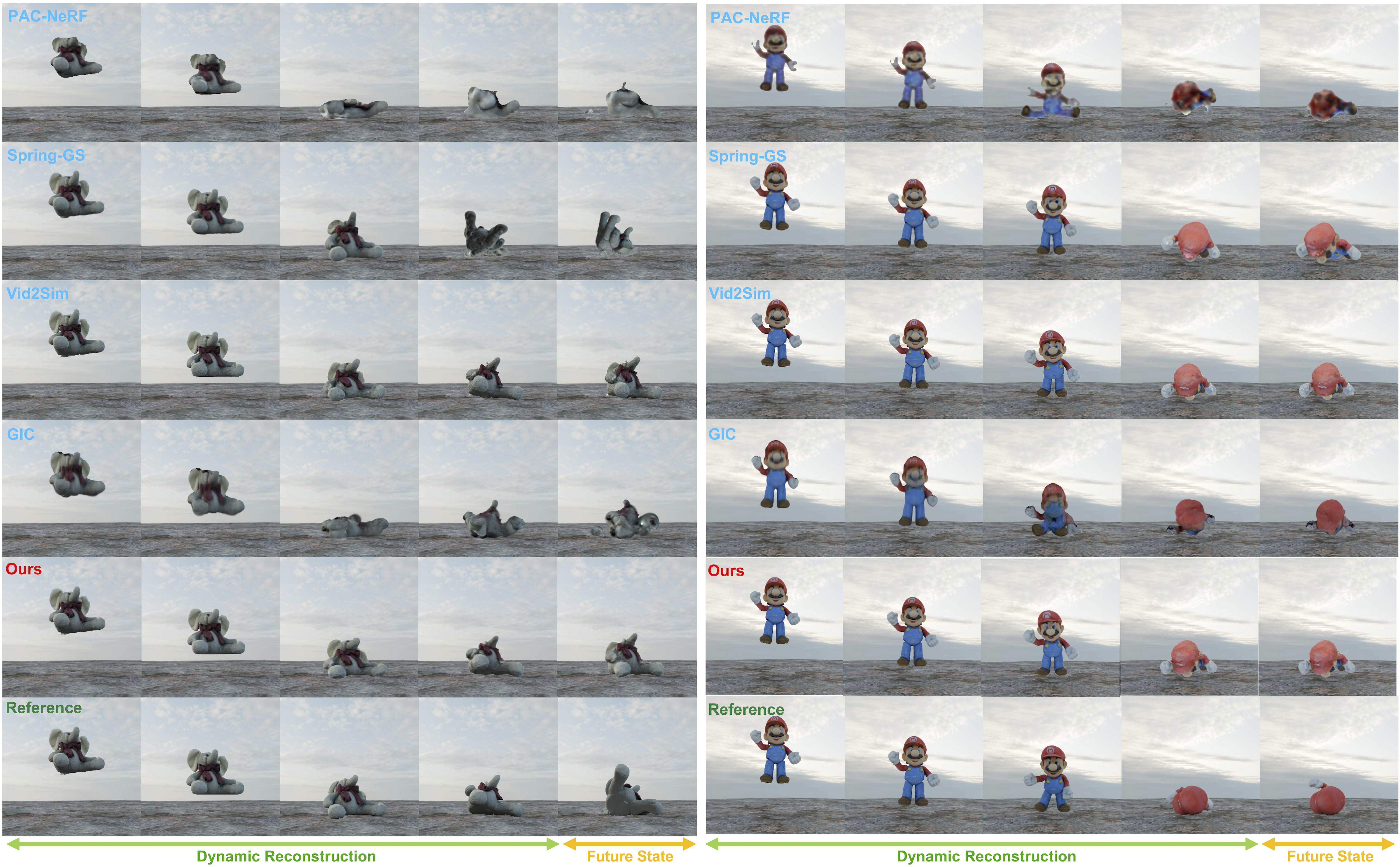} 
    \vspace{-5pt}
     \caption{\footnotesize Comparison with baseline~\cite{li2023pac,cai2024gic,zhong2024springgaus,chen2025vid2sim} on dynamic reconstruction and future state prediction. Our method preserves high-quality geometry and appearance while producing physically plausible dynamics.}
     \vspace{-5pt}
    \label{fig:dynamic}
\end{figure*}
\vspace{-5pt}
\subsection{Dynamic Reconstruction}
 \label{sec:dynamic}
%\vspace{-5pt}
Our method and baselines for dynamic reconstruction were evaluated using a set of 12 GSO test cases following \cite{li2023pac,zhong2024springgaus,cai2024gic,chen2025vid2sim}.  
For fair comparison, we use Vid2Sim's variant that trains 3D Gaussians from scratch without initializing 3D Gaussians from pre-trained Large Multi-view Gaussian Model (LGM)~\cite{tang2024lgm} predictions. 
As shown in Tab.~\ref{tab:dynamic} and Fig.~\ref{fig:dynamic}, PhysConvex achieves superior reconstruction quality and physical realism. Prior methods that optimize voxels in NeRF or ellipsoids in 3DGS often produce blurred textures or inaccurate dynamics, whereas our reduced-order, boundary-driven convex fields preserve sharp details and dynamic fidelity. Moreover, unlike grid-based MPM or spring-mass elastic dynamics, which are susceptible to numerical artifacts and oscillations in large-scale elastodynamics of complex geometries, our mesh- and grid-free reduced-space solver yields efficient, accurate dynamics.

\vspace{-5pt}
\begin{table*}[ht]
\centering
\vspace{-5pt}
\caption{\label{tab:dynamic} Quantitative Comparison on Dynamic Reconstruction (\textcolor{blue}{Best}).}
\vspace{-5pt}
\resizebox{\linewidth}{!}{
\begin{tabular}{c|c|cccccccccccc|c}
\toprule
\multicolumn{2}{c|}{} & backpack & bell & blocks & bus & cream & elephant & grandpa & leather & lion & mario & sofa & turtle & Mean $\uparrow$ \\ 
\midrule

\multirow{5}{*}{\rotatebox[origin=c]{90}{FoVVDP $\uparrow$}} & PAC-NeRF & 6.043 & 7.473 & 7.001 & 6.540 & 5.991 & 6.791 & 6.626 & 6.485 & 7.006 & 6.876 & 6.543 & 6.711 & 6.674 \\ 
& Spring-Gaus & 5.455 & 6.862 & 6.890 & 6.377 & 5.899 & 6.524 & 6.998 & 5.988 & 6.902 & 6.153 & 6.300 & 6.569 & 6.410 \\ 
& Vid2Sim w/o LGM & 7.778 & 6.884 & \textcolor{blue}{8.792} & 7.622 & \textcolor{blue}{9.117} & 7.468 & 6.246 & 8.949 & \textcolor{blue}{7.681} & 6.325 & \textcolor{blue}{8.034} & 8.628 & 7.794 \\
& GIC & 6.130 & 6.230 & 6.062 & 6.552 & 5.889 & 6.907 & 6.855 & 6.737 & 6.331 & 6.985 & 7.069 & 6.782 & 6.544 \\ 

& \cellcolor{blue!5} \textbf{Ours}    & \cellcolor{blue!5} \textcolor{blue}{8.403}    & \cellcolor{blue!5} \textcolor{blue}{7.843} & \cellcolor{blue!5} 8.783  & \cellcolor{blue!5} \textcolor{blue}{8.005} & \cellcolor{blue!5} 8.708 & \cellcolor{blue!5} \textcolor{blue}{8.178}    & \cellcolor{blue!5} \textcolor{blue}{7.752}   & \cellcolor{blue!5} \textcolor{blue}{9.153}   & \cellcolor{blue!5} 7.617 & \cellcolor{blue!5} \textcolor{blue}{7.792} & \cellcolor{blue!5} 7.925 & \cellcolor{blue!5} \textcolor{blue}{8.801}  & \cellcolor{blue!5} \textcolor{blue}{8.247}         \\
\midrule

\multirow{5}{*}{\rotatebox[origin=c]{90}{PSNR $\uparrow$}} & PAC-NeRF & 19.37 & 25.00 & 23.36 & 20.72 & 23.24 & 22.27 & 21.63 & 20.85 & 22.66 & 21.01 & 22.49 & 22.19 & 22.06 \\ 
& Spring-Gaus & 17.42 & 20.49 & 22.78 & 20.06 & 23.58 & 21.30 & 21.64 & 18.29 & 21.95 & 20.23 & 21.89 & 21.23 & 20.91 \\ 
& Vid2Sim w/o LGM & 26.87 & 26.59 & 32.57 & 26.53 & \textcolor{blue}{34.82} & 26.99 & 24.45 & 31.21 & \textcolor{blue}{27.62} & 24.83 & 30.01 & 30.33 & 28.57 \\
& GIC & 18.94 & 19.55 & 18.78 & 20.84 & 23.81 & 21.86 & 20.50 & 21.00 & 19.33 & 21.28 & 24.16 & 22.09 & 21.01 \\ 

& \cellcolor{blue!5} \textbf{Ours}    & \cellcolor{blue!5} \textcolor{blue}{28.09}    & \cellcolor{blue!5} \textcolor{blue}{29.67} & \cellcolor{blue!5} \textcolor{blue}{32.65}  & \cellcolor{blue!5} \textcolor{blue}{27.97} & \cellcolor{blue!5} 34.62 & \cellcolor{blue!5} \textcolor{blue}{29.17}    & \cellcolor{blue!5} \textcolor{blue}{27.67}   & \cellcolor{blue!5} \textcolor{blue}{33.50}   & \cellcolor{blue!5} 27.58 & \cellcolor{blue!5} \textcolor{blue}{27.46} & \cellcolor{blue!5} \textcolor{blue}{30.15} & \cellcolor{blue!5} \textcolor{blue}{31.49}  & \cellcolor{blue!5} \textcolor{blue}{30.00}         \\

\midrule
\multirow{5}{*}{\rotatebox[origin=c]{90}{SSIM $\uparrow$}} & PAC-NeRF & 0.887 & 0.956 & 0.940 & 0.908 & 0.893 & 0.922 & 0.939 & 0.932 & 0.936 & 0.921 & 0.926 & 0.923 & 0.924 \\ 
& Spring-Gaus & 0.867 & 0.941 & 0.941 & 0.903 & 0.912 & 0.919 & 0.948 & 0.917 & 0.937 & 0.920 & 0.921 & 0.920 & 0.920 \\ 
& Vid2Sim w/o LGM & 0.931 & 0.959 & 0.973 & 0.937 & 0.955 & 0.942 & 0.944 & 0.973 & 0.954 & 0.936 & 0.952 & 0.963 & 0.952 \\
& GIC & 0.903 & 0.945 & 0.930 & 0.925 & 0.922 & 0.936 & 0.948 & 0.949 & 0.934 & 0.938 & 0.942 & 0.936 & 0.934 \\ 

& \cellcolor{blue!5} \textbf{Ours}    & \cellcolor{blue!5} \textcolor{blue}{0.950}    & \cellcolor{blue!5} \textcolor{blue}{0.972} & \cellcolor{blue!5} \textcolor{blue}{0.975}  & \cellcolor{blue!5} \textcolor{blue}{0.948} & \cellcolor{blue!5} \textcolor{blue}{0.960} & \cellcolor{blue!5} \textcolor{blue}{0.957}    & \cellcolor{blue!5} \textcolor{blue}{ 0.964}   & \cellcolor{blue!5} \textcolor{blue}{0.980}   & \cellcolor{blue!5} \textcolor{blue}{0.958} & \cellcolor{blue!5} \textcolor{blue}{0.954} & \cellcolor{blue!5} \textcolor{blue}{0.955} & \cellcolor{blue!5} \textcolor{blue}{0.971}  & \cellcolor{blue!5} \textcolor{blue}{0.962}         \\

\bottomrule
\end{tabular}}
\end{table*}\vspace{-5pt}

\vspace{-5pt}
\subsection{Generalization Capability}
 \label{sec:generalization}
%\vspace{-5pt}
We further demonstrate the generalization of our method to novel materials, forces, and boundary conditions, showing consistent performance. 

\begin{figure*}[ht]
    \centering
    \includegraphics[width=1\textwidth]{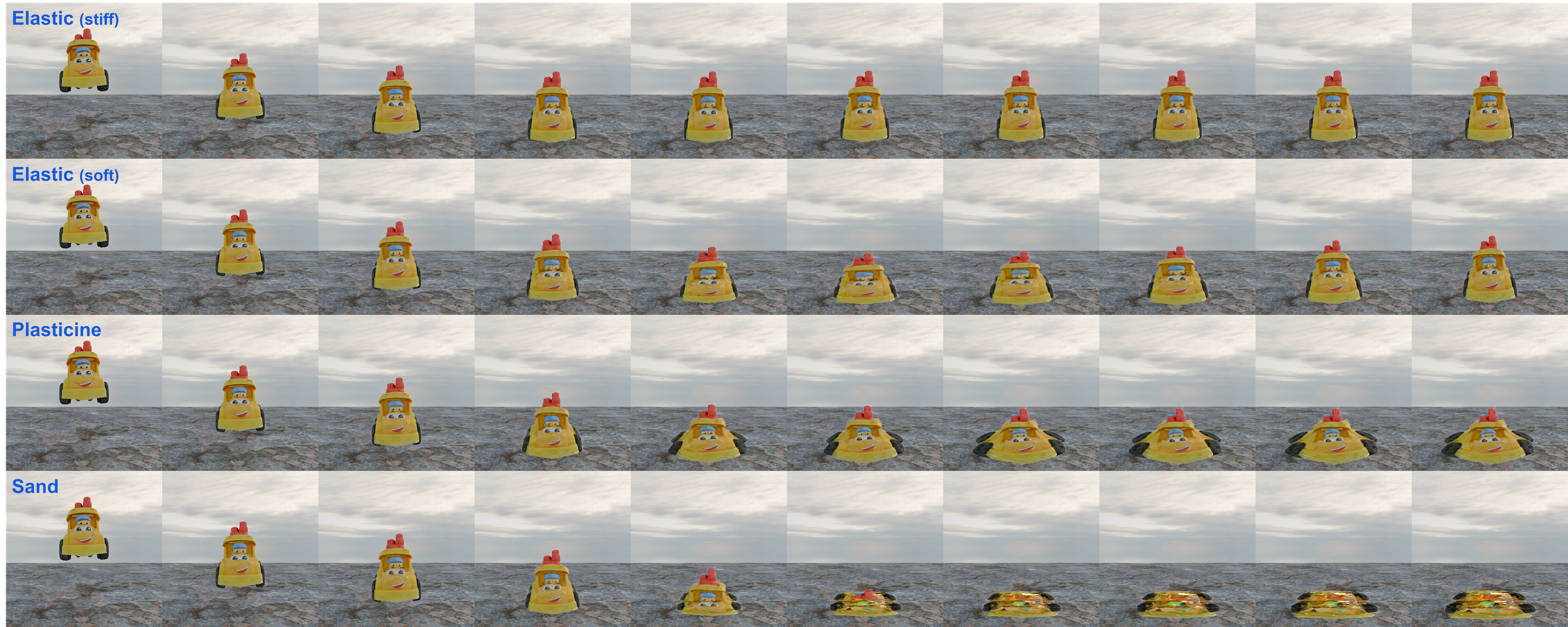} 
% \vspace{-5pt}
        \caption{\footnotesize Generalization to novel materials. Elastic (stiff): $E{=}10^7$, $\nu{=}0.49$. Elastic (soft): $E = 8000$, $\nu = 0.4$. Plasticine: $\tau_Y{=}500$. Sand: $\theta_f{=}10^\circ$.}
        %\vspace{-5pt}
    \label{fig:generalized:materials}
\end{figure*}

\textbf{\textit{Generalization to Different Materials.}}
PhysConvex generalizes naturally to different material behaviors and their constitutive models. We demonstrate this by simulating \textit{Elastic}, \textit{Plasticine}, and \textit{Sand} materials following ~\cite{li2023pac,cai2024gic}.
In Fig.~\ref{fig:generalized:materials}, our method captures the distinct material responses by utilizing corresponding energy densities derived for each model (see Supplementary). 

\begin{figure*}[ht]
    \centering
    \includegraphics[width=1\textwidth]{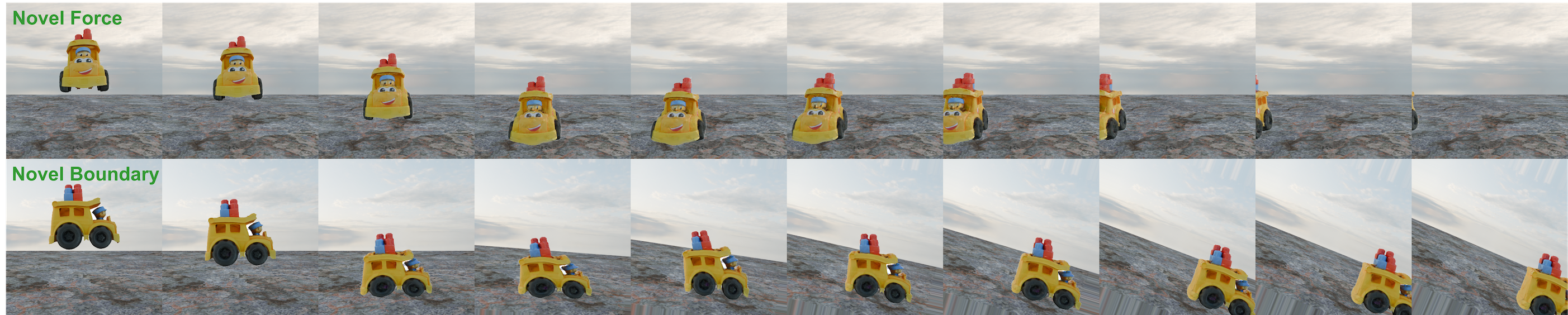} 
    %\vspace{-5pt}
    \caption{\footnotesize Generalization to novel force and complex boundary conditions. Force: A continuous external force pushing the falling object toward the left. Boundary: An object falls and is subjected to sliding motion on a continuously moving floor. }
    %\vspace{-5pt}
    \label{fig:generalized:boundary}
\end{figure*}

\textbf{\textit{Generalization to Novel Boundary and Force Conditions.}}
PhysConvex is robust to varying environmental dynamics, seamlessly integrating into dynamic scenes with novel forces and boundary interactions. As shown in Fig.~\ref{fig:generalized:boundary}, our method produces stable, high-quality reconstructions and physically consistent animations under diverse constraints. 

\vspace{-5pt}
\subsection{Future State Prediction}
 \label{sec:future}
%\vspace{-5pt}
We further evaluate temporal generalization following \cite{zhong2024springgaus}. After reconstructing the first 16 frames, all methods predict the subsequent 8 frames. Average results across all test cases (Tab.~\ref{tab:future} and Fig.~\ref{fig:dynamic}) show that PhysConvex maintains the top accuracy, demonstrating improved physical consistency in temporal evolution.

\noindent\begin{minipage}[t]{0.49\textwidth}
\captionsetup{type=table}
\caption{\label{tab:future} Future State Prediction.}
%\centering
\resizebox{.97\linewidth}{!}{
\begin{tabular}{c|ccccg}
\hline
& PAC-NeRF & Spring-Gaus & GIC   & Vid2Sim & \textbf{Ours}    \\\hline
PSNR$\uparrow$        & 20.11    & 18.32       & 19.20 & 25.07   & \textcolor{blue}{25.71}    \\
SSIM $\uparrow$       & 0.913    & 0.905       & 0.916 & 0.945   & \textcolor{blue}{0.948}    \\
FoVVDP $\uparrow$     & 5.948    & 5.443       & 5.702 & \textcolor{blue}{7.770}   & 7.683    \\
\hline
\end{tabular}
}
\end{minipage}\hfill
\begin{minipage}[t]{0.49\textwidth}
\captionsetup{type=table}
\caption{\label{tab:time} Training \& Primitive Efficiency.}
%\centering
\resizebox{.97\linewidth}{!}{
\begin{tabular}{c|ccccc}
\hline
               & PAC-NeRF & GIC    & Spring-GS & Vid2Sim & \our \textbf{Ours}  \\ \hline
Avg. Training  & \multirow{2}{*}{51}  & \multirow{2}{*}{69} & \multirow{2}{*}{32}    & \multirow{2}{*}{8}   & \multirow{2}{*}{\our\textcolor{blue}{6}} \\
Time (min) & & & & & \\
Primitive Num.  & -  &  - &  \textcolor{blue}{22882}  &  31363 & \our 23681 \\ \hline
\end{tabular}
}
\end{minipage}

\vspace{-5pt}
\subsection{Efficiency Comparison}
\label{sec:time}
%\vspace{-5pt}
We report the average total training time and the average number of primitives required for different methods using their default settings (excluding the primitives from network-based models, PAC-NeRF and GIC). As shown in Tab.~\ref{tab:time}, our method is notably more efficient, achieving shorter training time and requiring fewer primitives compared to even highly efficient baselines like Spring-Gaus and Vid2Sim. 
This superior efficiency stems from the compact representation achieved by our convex primitives and the inherent performance gains of our reduced-order simulation combined with an implicit Euler solver. 
All tests were executed on a single NVIDIA-RTX A6000 GPU.

\begin{table*}[!ht]
\caption{\label{tab:bounaryconvex:ablation} Ablation of Our Dynamic Convex Field with Boundary-driven Kinematics via Model Order Reduction Simulation (MOR). (\topa{Best},\topb{Second},\topc{Third})}
\centering
\resizebox{1\textwidth}{!}{
\begin{tabular}{c|cc|cc|c|ccc|cc}
\hline
\multirow{2}{*}{Model} 
& \multicolumn{2}{c|}{3D Primitive} 
& \multicolumn{2}{c|}{Kinematics}
& Simulation
& \multicolumn{3}{c|}{Dynamic Reconstruction}
& \multicolumn{2}{c}{Physics Identification} \\
\cline{2-11}
& Gaussian $\mid$ & Convex & Center $\mid$ & Boundary & MOR 
& PSNR$\uparrow$ & SSIM$\uparrow$ & FoVVDP$\uparrow$ 
& $\log(E)\downarrow$ & $\nu\downarrow$ \\
\hline
\textbf{A}                  & \checkmark &                           & \checkmark &                           & \checkmark & \third 28.65                        & \third 0.954                        & \third 7.895                        & {\second 0.54} & {\third 0.07}    \\
\textbf{B}                  &                           & \checkmark & \checkmark &                           & \checkmark & \second 29.16                        & \second 0.958                        & \second 8.173                        & \third 0.57                        & \second 0.05                         \\
\textbf{C}                  &                           & \checkmark &                           & \checkmark & \checkmark & {\best 30.00} & {\best 0.962} & {\best 8.247} & {\best 0.37} & {\best 0.04} \\\hline
\end{tabular}}
\end{table*}

\begin{table*}[htbp]
\caption{Ablation of $K$ of Convex Points ($M=10$). (\topa{Best},\topb{Second},\topc{Third}).} \label{tab:k:ablation}
\centering
\resizebox{1\textwidth}{!}{
\begin{tabular}{c|ccc|ccc|cc|c|c}
\midrule
\multicolumn{1}{l}{} & \multicolumn{3}{|c|}{Dynamic Reconstruction} & \multicolumn{3}{c|}{Future State Prediction} & \multicolumn{2}{c|}{Physics Identification} &    \multicolumn{1}{c|}{Primitives} &                           \multicolumn{1}{c}{Training Time}\\
\midrule
$K$              & PSNR $\uparrow$ & SSIM $\uparrow$ & FoVVDP $\uparrow$       & PSNR $\uparrow$ & SSIM $\uparrow$ & FoVVDP $\uparrow$        & $\log{(E)}$                   & $\nu$                        & Num.    &  Average   \\
\midrule
K=3             & 24.90        &  0.939        & 6.957        & 19.63         & 0.919         & 6.139         & 0.74                     & \second 0.05&\best 11827                          &\best 5min 20s\\
K=4             & 29.56        & \third 0.960& 8.151        & 24.50         & 0.944         & 7.500         & \third 0.38& \best 0.04                     & \second 14405&\second 5min 36s\\
K=5             & 29.88        & \second0.961& \third 8.235& \third 25.09& \second0.946& \second7.688&\best 0.36                     & \best0.04                     & \third 18773&\third6min 5s\\
K=6             &  \third 30.00        & \best 0.962        & \second8.247& \best 25.71& \best 0.948& \third 7.683& \second 0.37& \best 0.04& 23681                          &6min 14s\\
K=7             & \second 30.01        & \best 0.962        & \third 8.235& 24.95         & \third 0.945& 7.664         & \third 0.38& \second 0.05& 29646                          &6min 39s\\
K=8             & \best 30.19        &\best 0.962        &\best 8.290        & \second25.38&\best 0.948         &\best 7.760         & 0.47                     & \best0.04                     & 35320      &7min 5s\\ \hline               
\end{tabular}}
\end{table*}

\begin{table*}[!ht]
\caption{\label{tab:m:ablation} Ablation of the Number $M$ of Reduced DOFs. (\topa{Best},\topb{Second},\topc{Third})}
\centering
\resizebox{1\textwidth}{!}{
\begin{tabular}{c|c|ccc|ccc|cc|c}
\midrule   
\multicolumn{2}{l}{} & \multicolumn{3}{|c|}{Dynamic Reconstruction} & \multicolumn{3}{c|}{Future State Prediction} & \multicolumn{2}{c|}{Physics Identification} &    \multicolumn{1}{c}{Primitives}                             \\ \midrule 
K                    & M           & PSNR $\uparrow$ & SSIM $\uparrow$ & FoVVDP $\uparrow$       & PSNR $\uparrow$ & SSIM $\uparrow$ & FoVVDP $\uparrow$        & $\log{(E)}$                   & $\nu$                        & Num.\\
\midrule                   

\multirow{3}{*}{K=4} & M=8         & 29.47        & 0.959        & 8.152        & 24.55         & \third 0.944& \third 7.605& \third 0.351& 0.053                    & \second 14392\\
                     & M=10        & 29.56        & \third 0.960& 8.151        & 24.50         & \third 0.944& 7.500         & 0.376                    & \second 0.044& \third 14405\\
                     & M=12        & 29.60        & \third 0.960& 8.206        & 24.47         & 0.943         & 7.558         & 0.510                    & \best 0.043& \best 14180\\ \midrule   
\multirow{3}{*}{K=6} & M=8         & \second 29.96& \best 0.962& \second 8.214& \second 25.05& \second 0.946& \second 7.635& \second 0.330& \third 0.048& 23832                         \\
                     & M=10        & \best 30.00& \best 0.962& \best 8.247& \best 25.71& \best 0.948& \best 7.683& 0.371                    & \second 0.044& 23681                         \\
                     & M=12        & \third 29.88& \second0.961& \third 8.210& \third 24.76& \third 0.944& 7.601         & \best 0.321& \third 0.048& 23736   \\ \midrule                        
\end{tabular}}
\end{table*}

%\vspace{-5pt}
\subsection{Ablation Study}
\label{sec:ablation}
%\vspace{-5pt}

\textit{\textbf{Boundary-driven Kinematics Ablation.}}
We ablate the proposed boundary-driven kinematics compared with a center-driven convex variant. In Tab.~\ref{tab:bounaryconvex:ablation}, boundary-driven convex dynamics (Full Model \textbf{C}) consistently outperform center-driven convex dynamics (Model \textbf{B}), improving dynamic reconstruction and reducing physical parameter errors.
These results demonstrate that explicitly advecting convex boundaries, rather than only moving primitive centers, captures non-uniform deformation and yields more accurate physics reconstruction.

\textit{\textbf{Convex Representation Ablation.}}
To isolate the effect of the convex representation from the MOR simulation framework, we compare Gaussian and convex primitives under the same reduced-order simulation setting in Tab.~\ref{tab:bounaryconvex:ablation}.
Comparing the center-driven Gaussian variant (Model \textbf{A}) with the center-driven convex variant (Model \textbf{B}) shows that replacing Gaussians with convex primitives improves dynamic reconstruction and reduces the Poisson ratio error. This validates that convex primitives provide better volumetric coverage and geometric expressiveness than Gaussian primitives under the same reduced-order simulation. With the full boundary-driven convex formulation (Model \textbf{C}), the performance further improves, confirming the effectiveness of our dynamic convex field for both geometry and physics identification.

\textit{\textbf{DOFs $M$ and Convex Hull $K$ Ablation.}}
Ablation studies validate two key components: the number of 3D points $K$ per convex hull in our boundary-driven convex dynamics (Tab.~\ref{tab:k:ablation}) and the number of degrees of freedom (skinning handles) $M$ in our reduced-order convex simulation (Tab.~\ref{tab:m:ablation}). 
Increasing $K$ enhances the flexibility of convex topology representation and improves reconstruction quality, though it introduces higher computational cost and longer training time. However, the performance gain saturates beyond $K=6$. Similarly, increasing $M$ boosts deformation expressivity in simulation but is bounded by the capacity of the neural skinning network. 
Note that both $K$ and $M$ can be adjusted to accommodate more complex geometries and material deformations in different dynamic scenes.
Despite the added complexity, our convex representation remains highly efficient, requiring fewer primitives and shorter training time than Gaussian-based methods (see~Tab. \ref{tab:time}), owing to compact convex coverage and reduced-order simulation.

\begin{figure*}[ht]
    \centering
    \includegraphics[width=1\textwidth]{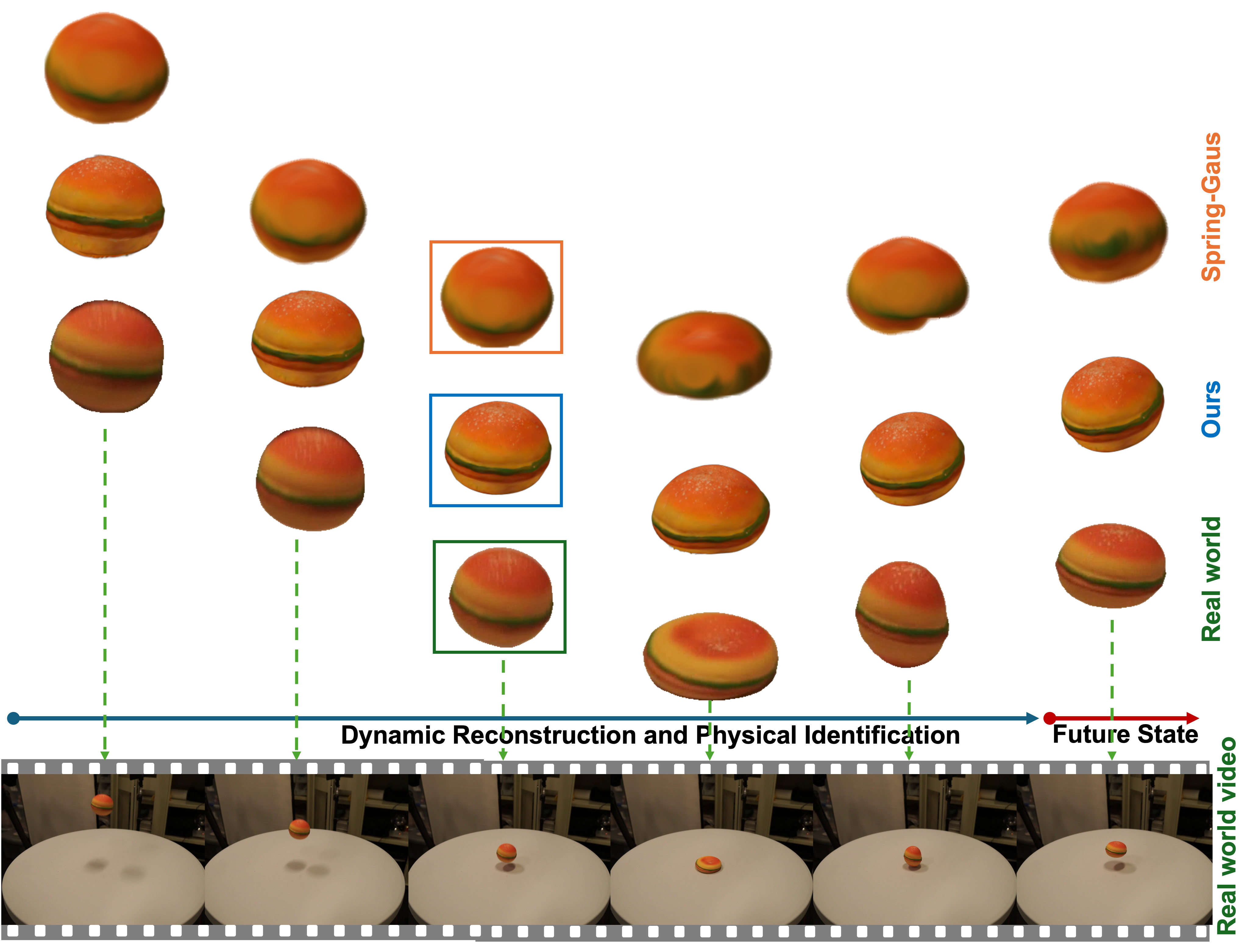} 
    
    \caption{\footnotesize Results on real-world samples (bottom: real references with background; top: zoomed-in objects without background). Compared with blurred real video observations and Spring-Gaus reconstructions, our method recovers geometry and appearance with clearer deformable details. Moreover, the reconstructed dynamics closely match the real video, exhibiting physically consistent motion and deformation.}
    \label{fig:real}

\end{figure*}

\vspace{-5pt}
\subsection{Real-world Evaluation}
\label{sec:realdata}
\vspace{-5pt}

Following Spring-Gaus~\cite{zhong2024springgaus}, we evaluate PhysConvex on real-world dynamic reconstruction and future state prediction, as shown in Fig.~\ref{fig:real}. 
Reconstructing and simulating dynamic scenes from sparse viewpoints and motion-blurred videos in real-world datasets is highly challenging. To address this, we adopt the registration network in~\cite{zhong2024springgaus} to align 3D convex kernels from static scene coordinates to dynamic scene coordinates. Furthermore, our reduced-order boundary-driven convex dynamics enable efficient and physically plausible dynamic reconstruction and simulation.
Compared with the blurred real-capture video and the reconstruction results of Spring-Gaus, our method recovers sharper geometry and appearance with finer details of deformable objects under real-world physics, such as the white dots on the burger object highlighted in the color boxes of Fig.~\ref{fig:real}. Moreover, our reconstructed dynamics closely match the real video, demonstrating physically consistent motion and deformation.

 \vspace{-5pt}
 \section{Conclusions and Future Work}
\label{sec:concl}
\vspace{-5pt}
We introduced PhysConvex, a physics-informed dynamic convex field for video-based reconstruction, physical system identification, and simulation of deformable objects. The key idea is to make each convex primitive a shared material-space unit for rendering, deformation, and physical simulation. By advecting convex boundaries rather than only primitive centers, PhysConvex captures non-uniform deformation, evolving active supports, sharp boundary motion, and compact anisotropic geometry.
We further introduced a mesh-free reduced-order convex simulator, where neural skinning modes define compact deformation bases over convex supports. The same supports are used to evaluate mass, elastic energy, forces, and contacts, avoiding dense grids, tetrahedral meshes, and predefined particle-primitive bindings. Coupled with differentiable convex splatting, PhysConvex jointly recovers appearance, geometry, dynamics, and material parameters from videos.
Experiments show improved dynamic reconstruction, physical parameter estimation, efficiency, future prediction, and generalization to changed materials, forces, and boundary conditions. 
Current limitations include the need for posed videos, object masks, and constrained physical settings such as gravity, boundary, and contact conditions. 
Future work will explore feed-forward convex initialization, contact modeling, and multi-object dynamic scene reconstruction.
%===============

\paragraph{\textbf{Acknowledgements.}}
This work was supported in part by ONR grant N00014-23-1-2526,  IARPA via DOI/IBC contract number 140D0423C0076, gifts from Adobe, Google and Qualcomm, an NSERC Postdoctoral Fellowship, the Ronald L. Graham Chair, and the UC San Diego Center for Visual Computing. The views and conclusions contained herein are those of the authors and should not be interpreted as necessarily representing IARPA, ONR, DOI/IBC, or the U.S. Government.

% \input{sec/06_appendix}
% \section*{Acknowledgements}
% Please insert your acknowledgments here.

\bibliographystyle{splncs04}
\bibliography{main}
 
\end{document}